\documentclass[conference]{IEEEtran}
\IEEEoverridecommandlockouts

\usepackage{cite}
\usepackage{amsmath,amssymb,amsfonts}
\usepackage{algorithmic}
\usepackage{graphicx}
\usepackage{textcomp}
\usepackage{xcolor}
\usepackage{booktabs}
\usepackage{multirow}
\usepackage{url}

\def\BibTeX{{\rm B\kern-.05em{\sc i\kern-.025em b}\kern-.08em
    T\kern-.1667em\lower.7ex\hbox{E}\kern-.125emX}}

\begin{document}

\title{Latent Cluster Separation as a Regularizer for Lifelong Anomaly Detection\\}

\author{
\IEEEauthorblockN{Sabarish Radhakrishnan}
\IEEEauthorblockA{John Champe High School / Academies of Loudoun\\
Aldie, VA, USA}
\and
\IEEEauthorblockN{Nathalie Japkowicz}
\IEEEauthorblockA{Department of Computer Science\\
American University\\
Washington, DC, USA}
\thanks{This work has been submitted to the IEEE for possible publication. Copyright may be transferred without notice, after which this version may no longer be accessible.}
}

\maketitle

\begin{abstract}
Lifelong (continual) anomaly detection systems must learn a sequence of tasks without forgetting how to recognize anomaly types seen earlier -- a property known as backward transfer. A common way to try to improve backward transfer is to reshape a model's internal representation around learnable concept centroids, typically coupled with a new way of scoring anomalies based on distance to those centroids. Because these two design choices are usually introduced together, it is unclear which one is actually responsible for any performance gain. In this paper, we unbundle the two mechanisms within a VAE-based lifelong anomaly detector and study them separately. We find that used purely as a training-time regularizer -- while leaving anomaly scoring unchanged -- latent cluster separation can improve detection quality and backward transfer, though whether it outperforms the alternative scoring approach depends on the dataset's underlying representation. Switching to distance-based scoring is instead where most of the instability commonly associated with this class of method actually originates; we identify and explain the underlying mechanism. This benefit depends on the dataset's underlying representation, and we report this as practical guidance for choosing between the two design choices, rather than as a list of caveats.
\end{abstract}


\section{Introduction}
\label{sec:intro}

Lifelong, or continual, learning systems must adapt to new tasks over time without forgetting what they have already learned. In anomaly detection, this means a lifelong detector should ideally improve at recognizing new anomaly types as it encounters them, without losing its ability to recognize types it saw earlier -- a property known as positive backward transfer. Achieving this reliably remains an open problem, particularly for detectors built on variational autoencoders (VAEs), where representation quality and the anomaly-scoring strategy are often tightly coupled.

One family of approaches tries to improve backward transfer by reshaping a VAE's latent space around learnable concept centroids. This is typically implemented as a single package: an attraction-repulsion loss reshapes the representation, and anomaly scoring switches from reconstruction error to distance-to-nearest-centroid to exploit that reshaped representation directly. This bundling obscures a simpler question: which half of the package is actually doing the work? In this paper, we unbundle the two design choices -- representation shaping and anomaly scoring -- inside VLAD's~\cite{vlad2023} VAE-based lifelong anomaly detector, and show that they behave very differently.

To see why this bundling is worth separating, consider the three latent-space configurations it conflates. In an \emph{unclustered} latent space (Baseline), an encoder places each input wherever its own features suggest, with no notion of task or concept; anomaly scoring simply measures how well the decoder reconstructs that point. In a \emph{clustered} latent space, an explicit training-time objective additionally pulls each task's points toward one of $K$ learned centroids and pushes centroids apart from each other, so that -- if it works as intended -- each concept the model has seen occupies its own distinguishable region. Once organized this way, two metrics can read it: \emph{distance-to-centroid} scoring, which asks how close a new point falls to the nearest known concept, or \emph{reconstruction-error} scoring, unchanged from the unclustered case. The intuition is straightforward: separating concepts into distinct regions should make it easier to remember a task even after later ones are learned, since new training has less opportunity to blur one concept's representation into another's -- exactly the backward-transfer property this paper is about. Whether this is actually the mechanism behind the regularizer's benefit, however, is a question this paper's own evidence complicates rather than confirms (Section~\ref{subsec:collapse}).

Bundling these two changes together also creates a specific coupling worth flagging before we test it: once scoring depends on distance to a centroid rather than reconstruction error, its quality is downstream of how well the representation-shaping objective actually organizes the latent space during training -- a dependency reconstruction-error scoring never has, since it reads directly off the input regardless of how clustered or unclustered the code happens to be. Intuitively, a \emph{better}-organized, more separated representation should make a distance-based signal easier to read, not harder, so instability tied specifically to this design would be a counterintuitive result, not an expected one. Whether that coupling actually produces the instability commonly attributed to this class of method, and under what conditions, is the open question we investigate.

We test this by isolating the two design choices in a controlled comparison, holding the training objective and centroid initialization fixed so that the only difference between conditions is how anomalies are scored at inference time. Used purely as a training-time regularizer, latent cluster separation reliably improves detection quality and backward transfer, with low variance and no early-task performance dip. Distance-based scoring, in contrast, is unstable under the same conditions, with substantially higher seed-to-seed variance despite sharing an identical training objective. A collapse diagnostic in the encoder's latent space explains the mechanism behind this gap: the shared representation can degenerate toward a near-constant code under either scoring rule, but reconstruction-based scoring remains informative despite this degeneration, while distance-based scoring does not.

This benefit depends on the dataset's underlying representation: for two of the three benchmarks we study, reconstruction-based scoring wins decisively on both accuracy and stability, while for the third, distance-based scoring is instead the better choice. We read this not as a failure to find a universal rule, but as the practical guidance the comparison makes possible: which scoring rule to pair with latent separation is a property of the data, not a free parameter to tune away.

As a byproduct of isolating these design choices, reconstruction-scoring regularization produces positive, low-variance backward transfer on ADFA-LD -- a property of pairing the separation objective with reconstruction-error scoring specifically, not of the objective alone, since switching only the scoring rule to distance-to-centroid, with everything else held fixed, is enough to erase it. Section~\ref{subsec:collapse}'s latent-space collapse diagnostic explains why.

Our contribution is threefold:
\begin{enumerate}
\item \textbf{Latent cluster separation, used as a training-time regularizer with reconstruction-error scoring left unchanged, reliably improves detection quality and backward transfer on ADFA-LD.} On ADFA-LD, LatentSep-Recon (latent separation used as a training-time regularizer) consistently beats Baseline on PR-AUC, achieves positive backward transfer, and shows no first-task cold-start penalty, with substantially lower variance than the distance-scoring alternative.
\item \textbf{The instability commonly associated with this class of method traces primarily to the distance-to-centroid scoring rule, not to the separation objective itself.} A controlled comparison holding the training objective and centroid initialization fixed isolates scoring as the sole variable between LatentSep-Recon and LatentSep-Dist (the distance-scoring variant), and a latent-space collapse diagnostic (Section~\ref{subsec:collapse}) shows the mechanism: both variants can collapse to a near-constant latent code, but only distance scoring loses its signal as a result, since reconstruction scoring still checks its output against the actual input.
\item \textbf{This benefit is dataset-dependent, and we isolate which design choice controls it.} Reconstruction-scoring regularization is the better choice on ADFA-LD and NSL-KDD; distance scoring is the better choice on UNSW-NB15, on PR-AUC. A mechanistic explanation for this split -- the collapse diagnostic in Section~\ref{subsec:collapse} -- shows why the scoring-rule variant is unstable while the regularizer variant is not.
\end{enumerate}

\section{Related Work}
\label{sec:related}

\textbf{Change-point-guided streaming anomaly detection.} CPDGA~\cite{cpdga2022} detects distributional change points in a data stream using a Wasserstein-distance-based statistical test introduced in~\cite{watch2021}, and uses them to decide when an anomaly detector should retrain, rather than retraining on a fixed schedule. This addresses plasticity directly, but the evaluation is forward-looking: how well the detector, once retrained, performs on the concept that triggered the retrain and on subsequent unseen concepts. Stability of previously learned concepts is not separately measured.

\textbf{Hierarchical memory and replay for lifelong anomaly detection.} VLAD~\cite{vlad2023} builds on the change-point-detection idea with a hierarchical memory (\texttt{HierarchicalLifewatchMemory}) that organizes previously encountered concepts and replays representative samples when retraining is triggered, so a new task does not overwrite the detector's ability to recognize older ones. Our work uses VLAD, unmodified in its change-point detection, memory organization, and replay logic, as the base system, changing only the anomaly detector's internal VAE model. Like CPDGA, VLAD's original evaluation does not report backward transfer.

\textbf{Positive backward transfer in continual classification.} Outside anomaly detection, positive backward transfer has been explicitly pursued and demonstrated in continual classification. CUBER~\cite{cuber2022} and LEAF~\cite{leaf2024} both report settings in which accuracy on earlier classes improves, rather than merely degrades less, as new classes are learned. These results motivated our initial question: can a mechanism in the anomaly-detection setting produce the same effect? Anomaly-detection work has since shown it can -- CITADEL~\cite{citadel2025} and CND-IDS~\cite{cndids2025}, discussed below, both report positive backward transfer directly, at larger magnitude than we do here -- which sharpens rather than closes the question we pursue: not whether positive BWT is achievable in continual anomaly detection, but which specific design choice inside a latent-separation mechanism produces it reliably, versus destroys it, and why. As Section~\ref{sec:results} shows, our answer isolates that choice precisely, rather than offering a universal claim about the mechanism as a whole.

\textbf{Latent-space separation for anomaly detection.} A separate line of work uses variational latent-space clustering or prototype separation to improve anomaly detection quality, under the intuition that novelty is better captured by distance in a well-separated latent space than by a single global reconstruction-error threshold (CL-VAE-style approaches)~\cite{clvae}. Our method is a direct, minimal instantiation of this idea inside VLAD's existing VAE component, adding only a term that organizes the encoder's latent space around learnable concept centroids. Unlike prior work here, we separately evaluate whether the resulting structure is useful only as a training-time regularizer or also as the scoring mechanism itself.

\section{Method}
\label{sec:method}

\subsection{Base system}
We use VLAD~\cite{vlad2023} unmodified as our base lifelong anomaly detection system. VLAD processes a sequence of tasks (concepts) $T_1, \dots, T_n$ one at a time under an \texttt{IncrementalTaskLearner} strategy: for each task, the current model is evaluated on every task's held-out test set (producing the full task-by-task evaluation matrix used for our metrics, Section~\ref{sec:setup}), then the model is updated on the new task's training data. A Wasserstein-distance-based change-point detector monitors the incoming stream and, together with a fixed step budget (\texttt{retrain\_after\_steps}=1000), decides when the underlying anomaly-scoring model should be retrained. When retraining is triggered, VLAD's hierarchical memory (\texttt{HierarchicalLifewatchMemory}) organizes previously seen concepts and supplies a replay buffer of representative samples from earlier concepts, which is combined with the new task's data for retraining. The change-point detector's sensitivity is controlled by a \texttt{threshold\_ratio} parameter that scales a per-dataset Wasserstein-distance threshold; VLAD's original codebase fixes this at a default of 2.0, a value under which none of our three datasets ever splits memory beyond a single concept (Section~\ref{sec:setup}). None of this -- change-point detection, memory organization, or replay -- is modified in this paper; we do, however, treat \texttt{threshold\_ratio} as a tunable experimental setting rather than a fixed default, applying the same value to both models under comparison so the comparison stays fair.

\subsection{Baseline VAE}
The anomaly-scoring component inside VLAD is a variational autoencoder. An encoder maps an input $x$ to a latent mean $\mu(x)$ and log-variance $\log\sigma^2(x)$ via a single hidden \texttt{Dense} layer; a latent sample $z$ is drawn via the standard reparameterization trick $z = \mu(x) + \sigma(x)\odot\epsilon,\ \epsilon\sim\mathcal{N}(0, I)$; and a decoder reconstructs $\hat{x}$ from $z$ through a mirrored \texttt{Dense} architecture with a sigmoid output layer. Both the encoder's and decoder's hidden layers have width $\max(8, \lfloor d/8 \rfloor)$ and the latent dimension is $\max(4, \lfloor d/32 \rfloor)$, where $d$ is the input dimensionality; for all three datasets here ($d=6, 41, 43$) both floors bind, giving a hidden width of 8 and a latent dimension of 4 uniformly. The model is trained with the standard VAE objective, reconstruction error plus KL divergence to a standard normal prior:
\begin{equation}
\mathcal{L}_{\text{VAE}} = \sum_i (x_i - \hat{x}_i)^2 \;-\; \tfrac{1}{2}\sum_j\left(1 + \log\sigma_j^2 - \mu_j^2 - \sigma_j^2\right).
\label{eq:vae}
\end{equation}
Training uses the Adam optimizer (learning rate $0.001$, gradients clipped to $[-0.5, 0.5]$), batch size 32, and 64 epochs per retraining event. At inference time, an input's anomaly score is its reconstruction mean-squared error, and the decision threshold is the maximum reconstruction error observed on the most recent training data (a max-threshold rule consistent with VLAD's original design). We refer to this unmodified scoring model as \textbf{Baseline}.

\subsection{Latent cluster-separation VAE}
\label{sec:method-latentsep}
Our modification changes two things, both confined to the VAE model file and invisible to the rest of the VLAD pipeline. We treat both as independently configurable, because Section~\ref{sec:results} evaluates their contributions separately.

\textbf{(a) Learnable latent centroids with an attraction-repulsion loss.} We introduce $K{=}8$ learnable centroid vectors $c_1, \dots, c_K$ in the latent space, trained jointly with the encoder/decoder weights via a custom gradient-tape training loop (needed because the centroids are not part of the standard Keras \texttt{Model} graph). For a batch of latent means $\{\mu(x_i)\}$, the separation loss combines an attraction term, which pulls each sample toward its nearest centroid,
\begin{equation}
\mathcal{L}_{\text{attract}} = \frac{1}{N}\sum_i \min_k \lVert \mu(x_i) - c_k \rVert_2^2,
\label{eq:attract}
\end{equation}
with a margin-based repulsion term (margin $m{=}3.0$) that keeps centroids from collapsing together,
\begin{equation}
\mathcal{L}_{\text{repel}} = \frac{1}{K(K-1)}\sum_{k \ne k'} \max(0,\, m - \lVert c_k - c_{k'} \rVert_2)^2.
\label{eq:repel}
\end{equation}
The total training loss is $\mathcal{L} = \mathcal{L}_{\text{VAE}} + \lambda\,(\mathcal{L}_{\text{attract}} + \mathcal{L}_{\text{repel}})$, with separation weight $\lambda{=}50$, chosen so the separation term visibly moves the centroids without destabilizing the base VAE loss. Centroids are initialized either randomly ($\mathcal{N}(0, 1)$ in each latent dimension, the original design) or, in a second variant we test, via $k$-means fit on the first task's encoder-mean output before any encoder training has occurred (\texttt{centroid\_init=kmeans}); in both cases centroids are re-optimized, not reset, at every subsequent retraining event. We fix $K{=}8$ throughout rather than sweeping it; \texttt{threshold\_ratio}, whose effect on results we examine directly (Section~\ref{sec:setup-tuning}), is the more consequential setting, and a finer, per-seed sweep of it is left to future work.

\textbf{(b) Anomaly scoring: reconstruction error or distance-to-centroid.} The separation objective in (a) can be paired with either of two scoring rules at inference time. \emph{Distance scoring}, as originally proposed, computes an input's anomaly score as its Euclidean distance, in latent space, to the nearest of the $K$ centroids, using the encoder's mean output $\mu(x)$ (not a stochastic sample):
\begin{equation}
s(x) = \min_k \lVert \mu(x) - c_k \rVert_2.
\end{equation}
\emph{Reconstruction scoring} instead keeps the baseline's reconstruction-error score unchanged -- the first term of Equation~\eqref{eq:vae}, $\sum_i (x_i - \hat{x}_i)^2$ -- using the separation objective only to shape the encoder's representation during training and not to determine how anomalies are scored at inference time. In both cases, the decision threshold is, symmetrically with the baseline, the maximum score observed on the most recent training data.

We refer to the three resulting configurations, all sharing identical change-point detection, memory, and replay logic, as: \textbf{Baseline} (no centroids, reconstruction scoring), \textbf{LatentSep-Recon} (centroids and separation loss trained, reconstruction scoring), and \textbf{LatentSep-Dist} (centroids and separation loss trained, distance scoring) -- the last of these being the more commonly implemented design. Section~\ref{sec:results} evaluates all three configurations and separately summarizes Baseline vs.\ LatentSep-Dist under one fixed configuration.

\textbf{What is not changed.} Change-point detection, hierarchical memory organization and summarization, replay sample selection, the retraining schedule, and the surrounding \texttt{IncrementalTaskLearner} strategy -- all inherited unmodified from VLAD's base architecture (Section~\ref{sec:method}) -- are identical across all three configurations above. All three also share the same input min-max normalization (fit per retraining event on the current training data) and the same numerical-stability safeguards described in Section~\ref{sec:setup}, and, within any single comparison, share the same \texttt{threshold\_ratio}, so that the comparisons in Section~\ref{sec:results} isolate the effect of the specific design choice under test.

\section{Experimental Setup}
\label{sec:setup}

\subsection{Datasets}
We test our three configurations on three public intrusion-detection benchmarks spanning two feature representations, to determine whether any observed effect is representation-specific: syscall-based (ADFA-LD) and hand-engineered connection/flow statistics (NSL-KDD, UNSW-NB15).

\textbf{ADFA-LD}~\cite{adfa2013}\textbf{.} Each task is one of 6 attack types (Adduser, Hydra\_FTP, Hydra\_SSH, Java\_Meterpreter, Meterpreter, Web\_Shell). Inputs are length-6 sliding windows of raw Linux system-call IDs (6 features, integer-valued, range approximately 0--340). For each task, training data is a stream of normal-only syscall windows; the test set combines held-out normal windows with that attack type's windows, labeled 0/1. We use a stratified subsample (5{,}000 training / 10{,}000 test rows per task, fixed seed 42, original normal:attack ratio preserved) for tractable experiment turnaround, since the unmodified Wasserstein-distance change-point detector does not scale to the full dataset in reasonable wall-clock time.

\textbf{NSL-KDD}~\cite{nslkdd2009}\textbf{.} Each task is one of the 4 standard top-level attack categories (DoS, Probe, R2L, U2R), each an aggregation of $\sim$10--20 specific attack labels present in the raw \texttt{KDDTrain+}/\texttt{KDDTest+} files. Inputs have 41 features: 3 categorical (protocol type, service, flag), label-encoded to integers, and 38 numeric connection/traffic statistics. Several numeric columns (\texttt{src\_bytes}, \texttt{dst\_bytes}, \texttt{duration}, and other count-valued columns) are extremely heavy-tailed (e.g., \texttt{src\_bytes} ranges from 0 to $\sim$89.5 million, with a 99th percentile of only $\sim$72 thousand); we apply a $\log(1+x)$ transform to these columns before scaling, standard practice for this dataset, to keep per-task min-max normalization numerically stable. As with ADFA-LD, each task's training data is that task's share of normal-labeled connections, and its test set combines held-out normal connections with all test-set connections in that attack category. Test-set sizes are bounded by natural availability rather than fixed at 10{,}000: NSL-KDD's U2R category, in particular, has only 67 attack examples in the entire test file.

\textbf{UNSW-NB15}~\cite{unsw2015}\textbf{.} Each task is one of 9 attack categories (Analysis, Backdoor, DoS, Exploits, Fuzzers, Generic, Reconnaissance, Shellcode, Worms), giving a substantially longer task sequence than the other two benchmarks. Inputs have 43 features: 3 categorical (protocol, service, connection state), label-encoded to integers, and 40 numeric connection-level statistics. As with NSL-KDD, several numeric columns are extremely heavy-tailed (e.g., \texttt{sload} ranges up to $\sim$6 billion, \texttt{sbytes}/\texttt{dbytes} up to $\sim$13--15 million), so we apply the same $\log(1{+}x)$ transform before scaling. Each task's training data is a round-robin share of a held-out pool of normal-labeled rows; its test set combines a disjoint round-robin share of held-out normal rows with all rows of that attack category. We use the same stratified-subsampling convention as ADFA-LD (5{,}000 training / 10{,}000 test rows per task, fixed seed) for tractable turnaround.

\subsection{A numerical stability fix, applied identically to all configurations}
NSL-KDD's heavy-tailed columns initially produced NaN anomaly scores by overflowing the reparameterization step's exponential when a task's normalization scale was small; we fixed this with the $\log(1{+}x)$ transform described above (applied to both NSL-KDD and UNSW-NB15) and a numerical-safety clip of $\log\sigma^2 \in [-10, 10]$ before the exponential in the sampling step. Both fixes are applied identically to code shared across all configurations, and we verified zero NaN scores across all (task-learned $\times$ evaluation-task) combinations before running the experiments reported below.

\subsection{Change-point threshold and centroid-initialization tuning}
\label{sec:setup-tuning}
At VLAD's default \texttt{threshold\_ratio}=2.0, ADFA-LD never splits memory beyond a single concept (Table~\ref{tab:ablation}, condition (a)); NSL-KDD and UNSW-NB15 show the same behavior at this default, so the hierarchical-memory and replay machinery motivating VLAD's design is never exercised on any of the three datasets. \texttt{threshold\_ratio}=1.0 was selected on ADFA-LD for producing a stable, non-degenerate concept count (2--6). We apply \texttt{threshold\_ratio}=1.0 to \emph{both} models on all three datasets (Section~\ref{sec:results}), since it is a property of the shared change-point/memory component, not either VAE variant. This value was tuned on ADFA-LD only and does \emph{not} transfer uniformly: at 1.0, ADFA-LD splits into 3 concepts, UNSW-NB15 into 2, NSL-KDD does not split at all. We report this as a limitation, not a generalizable recipe -- \texttt{threshold\_ratio} scales a threshold computed from each dataset's own Wasserstein-distance scale, so a value tuned on one representation should need re-tuning on another.

For centroid initialization, we test both the original random ($\mathcal{N}(0,1)$) scheme and a $k$-means alternative (Section~\ref{sec:method-latentsep}) intended to give the separation objective a better-than-random starting point and thereby reduce a cold-start effect. Unless noted otherwise, results reported for LatentSep-Dist in Section~\ref{sec:results} use $k$-means initialization together with \texttt{threshold\_ratio}=1.0 (the "tuned configuration"); Section~\ref{sec:discussion} later reports that this combination does not reduce cold-start variance.

\subsection{Metrics}
Following each task's training step, the current model is evaluated on \emph{every} task's test set, producing a task-by-task performance matrix for both PR-AUC and ROC-AUC. From this matrix we report, as is standard in continual learning evaluation:
\begin{itemize}
\item \textbf{Overall value}: mean performance across the diagonal-and-later entries of the matrix (performance on each task after it, and any task after it, has been learned).
\item \textbf{Backward transfer (BWT)}~\cite{gem2017}: the mean change in a task's performance between when it was first learned and after all subsequent tasks have been learned. Positive BWT means later learning improved earlier-task performance. Precisely,
\begin{equation}
\text{BWT} = \frac{1}{N-1}\sum_{i=1}^{N-1} \left(R_{i,N} - R_{i,i}\right),
\label{eq:bwt}
\end{equation}
where $R_{i,j}$ is performance on task $i$'s test set after training through task $j$ ($j \ge i$), $N$ is the total number of tasks, $R_{i,i}$ is task $i$'s performance immediately after it was first learned, and $R_{i,N}$ is its performance after all $N$ tasks have been learned.
\end{itemize}
We run every configuration with 3 random seeds (42, 123, 456); \texttt{random}, \texttt{numpy}, and \texttt{tensorflow} global seeds are set at the start of the script, before the data reader, model, and memory are constructed, so that centroid initialization, encoder/decoder weight initialization, and any batch-level stochasticity are seeded identically per seed. Standard deviations are computed with Bessel's correction ($n{-}1$ denominator, i.e.\ sample rather than population standard deviation) throughout.

\section{Results}
\label{sec:results}

\subsection{Baseline vs.\ LatentSep-Dist: A Direct Cross-Dataset Comparison}
\label{subsec:supporting}
We ran a direct Baseline-vs-LatentSep-Dist comparison -- under one fixed ``tuned'' configuration (\texttt{threshold\_ratio}=1.0, $k$-means initialization) -- on each dataset. This comparison conflates two simultaneous changes (whether centroids and the separation loss are trained at all, and which scoring rule is used), so while it shows the raw effect of adopting the full method, it cannot on its own attribute that effect to either change individually; Section~\ref{subsec:ablation}'s single-factor comparison isolates the scoring rule specifically and resolves that ambiguity. This comparison is deliberately two-condition: it asks whether adopting the full method -- centroids, separation loss, and distance scoring together -- is worth it relative to not changing anything, not which design choice is responsible. LatentSep-Recon, built to answer that second question, appears only in Section~\ref{subsec:ablation}.

\begin{table}[!t]
\caption{Baseline vs.\ LatentSep-Dist, tuned configuration (\texttt{threshold\_ratio}=1.0, $k$-means init): per-seed PR-AUC across all three datasets}
\label{tab:tuned-perseed}
\centering
\begin{tabular}{@{}lrr@{}}
\toprule
Dataset & Baseline & LatentSep-Dist \\
\midrule
ADFA-LD (seed 42)      & 0.1539 & 0.1350 \\
ADFA-LD (seed 123)     & 0.1502 & 0.1755 \\
ADFA-LD (seed 456)     & 0.1507 & 0.1228 \\
ADFA-LD (mean$\pm$std) & 0.1516$\pm$0.0020 & 0.1445$\pm$0.0276 \\
\addlinespace
NSL-KDD (seed 42)      & 0.8710 & 0.8167 \\
NSL-KDD (seed 123)     & 0.8717 & 0.8120 \\
NSL-KDD (seed 456)     & 0.8717 & 0.7841 \\
NSL-KDD (mean$\pm$std) & 0.8715$\pm$0.0004 & 0.8043$\pm$0.0176 \\
\addlinespace
UNSW-NB15 (seed 42)      & 0.9368 & 0.9608 \\
UNSW-NB15 (seed 123)     & 0.9462 & 0.9047 \\
UNSW-NB15 (seed 456)     & 0.9487 & 0.9461 \\
UNSW-NB15 (mean$\pm$std) & 0.9439$\pm$0.0063 & 0.9372$\pm$0.0291 \\
\bottomrule
\end{tabular}
\end{table}

Table~\ref{tab:tuned-perseed} shows per-seed detail for PR-AUC specifically; PR-AUC BWT, ROC-AUC, and ROC-AUC BWT for this same comparison are reported as seed-win counts (Table~\ref{tab:summary}). Table~\ref{tab:summary} shows that no metric wins on a majority of seeds on more than one dataset simultaneously, and NSL-KDD does not favor LatentSep-Dist on any metric tested. One pattern from Table~\ref{tab:tuned-perseed} holds without exception across every dataset and every metric: LatentSep-Dist's seed-to-seed standard deviation is larger than Baseline's -- ranging from roughly $2\times$ (UNSW-NB15 PR-AUC-BWT: Baseline std 0.0031 vs.\ LatentSep-Dist std 0.0067) to nearly $60\times$ (NSL-KDD ROC-AUC: Baseline std 0.0007 vs.\ LatentSep-Dist std 0.0409) larger, depending on dataset and metric. Section~\ref{subsec:ablation}'s single-factor comparison shows that once the scoring rule is fixed to reconstruction error, this variance problem resolves favorably on two of the three datasets (ADFA-LD, NSL-KDD) and unfavorably on the third (UNSW-NB15) -- a dataset-dependent fix, not a universal one.

\begin{table}[!t]
\caption{Cross-Dataset Summary: LatentSep-Dist Wins Against Baseline, in Seeds Out of 3, Tuned Configuration}
\label{tab:summary}
\centering
\begin{tabular}{@{}lccc@{}}
\toprule
Metric & ADFA-LD & NSL-KDD & UNSW-NB15 \\
\midrule
PR-AUC (overall)  & 1/3 & 0/3 & 1/3 \\
ROC-AUC (overall) & 1/3 & 0/3 & 2/3 \\
PR-AUC BWT        & 2/3 & 0/3 & 1/3 \\
ROC-AUC BWT       & 2/3 & 0/3 & 1/3 \\
\bottomrule
\end{tabular}
\end{table}

\subsection{Mechanism: A Latent-Space Collapse Diagnostic}
\label{subsec:collapse}

\begin{figure*}[!t]
\centering
\includegraphics[width=0.95\textwidth]{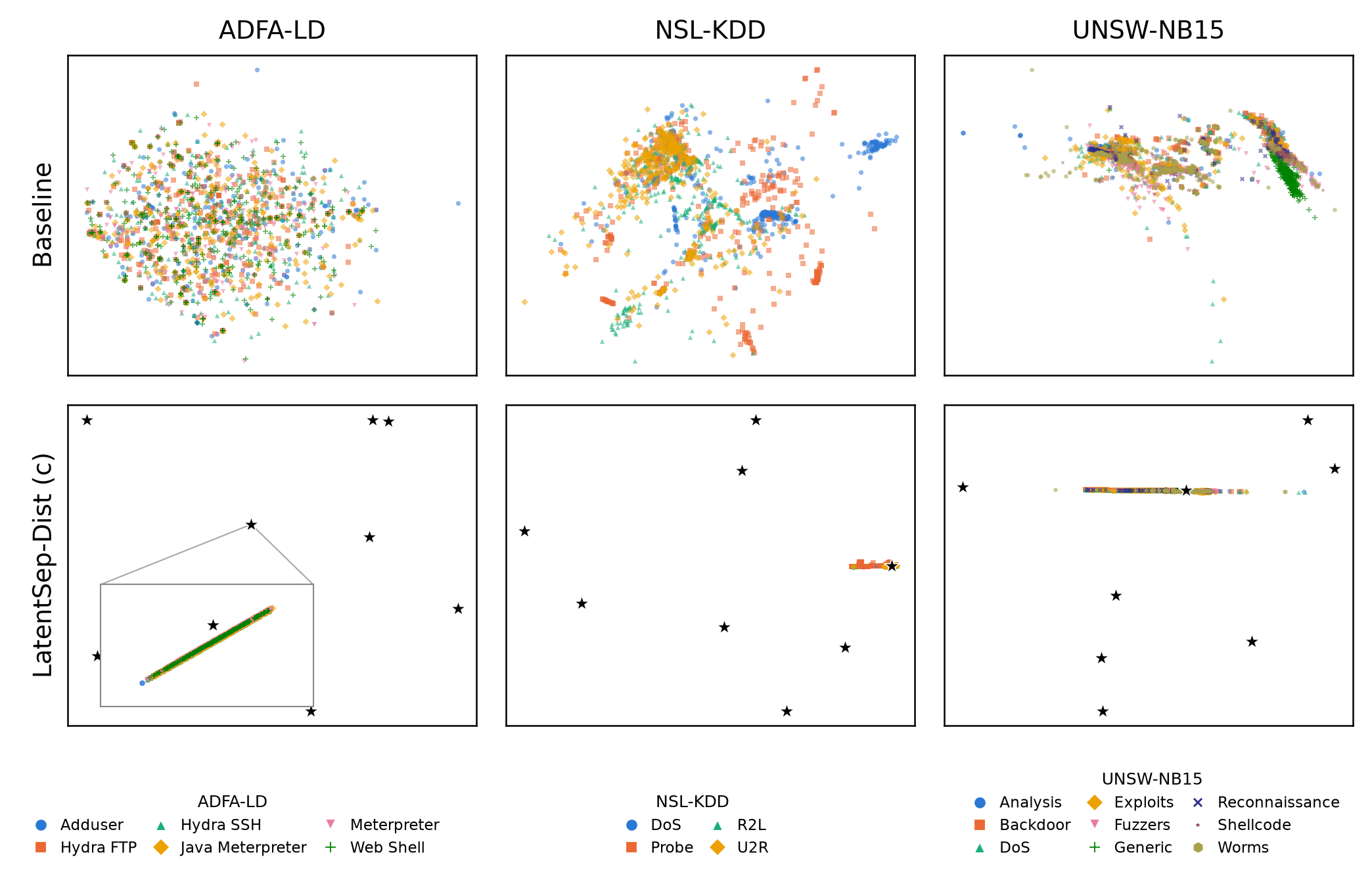}
\caption{Latent space (PCA projection), all three datasets, Baseline (top) and LatentSep-Dist (bottom), seed 42. LatentSep-Dist's points collapse onto one of the 8 trained centroids in essentially every case across all three datasets (98--100\% of each task's points, exactly 100\% for ADFA-LD and 8 of UNSW-NB15's 9 tasks; ADFA-LD's collapse, the tightest of the three, is shown via inset zoom); the other 7 centroids spread out normally. Baseline shows no such collapse in any dataset.}
\label{fig:latent-collapse}
\end{figure*}

To understand \emph{why} LatentSep-Dist is so much more variable than Baseline (Section~\ref{subsec:supporting}), we visualized the trained encoder's latent space across all three datasets in both Baseline and LatentSep-Dist conditions -- six runs total, one seed (42) each -- projecting $z_{\text{mean}}$ for 400 test points/task (2400 points for ADFA-LD, 1600 for NSL-KDD, 3600 for UNSW-NB15) and, for LatentSep-Dist, the 8 trained centroids, to 2D via PCA (Figure~\ref{fig:latent-collapse}).

Two findings emerge: one universal across all three datasets, one that is not.

\textbf{Collapse signature (cross-dataset).} Under LatentSep-Dist, points collapse onto a single dominant centroid regardless of dataset: ADFA-LD's 6 tasks and 8 of UNSW-NB15's 9 tasks assign 100\% of their points to one centroid; UNSW-NB15's ninth task (Worms) assigns 399/400 (99.75\%); NSL-KDD's 4 tasks assign 98--99\%. The unused centroids remain well-separated from each other in every dataset (pairwise distances exceed the $m{=}3.0$ training margin throughout), so the centroids themselves are not degenerate -- only the encoder's routing of data onto them is.

\textbf{Why the loss structure permits this.} The attraction term (Eq.~\eqref{eq:attract}) rewards each sample only for being close to its \emph{nearest} centroid, with no term rewarding balanced usage across centroids: attraction is minimized just as well if every sample's nearest centroid is the same one as if samples were spread across all $K$. The repulsion term (Eq.~\eqref{eq:repel}) acts purely between centroids' own positions, independent of whether any centroid actually attracts data. Together, these two terms impose no cost on a solution where one centroid absorbs the entire batch while the remaining $K-1$ satisfy repulsion among themselves without ever being assigned a sample -- exactly the winner-take-all pattern observed above. Ruling this out structurally could require a term rewarding balanced centroid usage (e.g., an entropy bonus on the empirical assignment distribution); the loss as designed does not include one.

\textbf{Collapse magnitude (dataset-dependent).} How tightly points cluster around that centroid, relative to the same encoder's latent-space spread under Baseline, varies by about two orders of magnitude across datasets. Taking each task's median distance from its own (outlier-robust) center under Baseline and its median distance to the nearest centroid under LatentSep-Dist, the Baseline-to-LatentSep-Dist ratio is 460.7--606.4$\times$ on ADFA-LD, 28.2--49.8$\times$ on NSL-KDD, and 3.8--10.5$\times$ on UNSW-NB15 ($\log_{10}(606.4/3.8) \approx 2.2$). ADFA-LD collapses nearly to a point; UNSW-NB15 converges onto the correct centroid but stays comparatively diffuse around it (visible in Figure~\ref{fig:latent-collapse} as a band, not a point); NSL-KDD sits between the two.

\textbf{A separate, NSL-KDD-specific artifact.} Independently of collapse, NSL-KDD alone shows a small fraction of test points (16/1600, 1.0\%, identical under both Baseline and LatentSep-Dist since the cause is the data/normalization pipeline, not the VAE variant) with $z_{\text{mean}}$ magnitude up to six orders larger than the rest of that dataset's latent space. We traced this to four columns (\texttt{land}, \texttt{wrong\_fragment}, \texttt{urgent}, \texttt{is\_host\_login}) that sit at exactly 0 for essentially every row of every task's 5{,}000-row training slice; a zero-variance training column collapses per-task min-max normalization scale to the numerical floor, so any rare nonzero occurrence of that feature elsewhere is amplified by the unclipped, linear $z_{\text{mean}}$ layer ($\log\sigma^2$ is clipped, Section~\ref{sec:setup}; $z_{\text{mean}}$ is not). This is not a general heavy-tail effect: UNSW-NB15's heavy-tailed columns (\texttt{sload}, \texttt{sbytes}, \texttt{dur}) are continuous flow statistics, never zero-variance within any task's slice, and UNSW-NB15 shows zero such outliers (0/3600) despite larger raw magnitudes than NSL-KDD (\texttt{sload} up to $\sim$6 billion vs.\ \texttt{src\_bytes}'s $\sim$89.5 million, Section~\ref{sec:setup}) -- the artifact tracks rare-event-flag column semantics, not tail size. These points are excluded only from the PCA views in Figure~\ref{fig:latent-collapse}; every statistic above is computed on the full, unfiltered data.

LatentSep-Recon at the same seed collapses identically to LatentSep-Dist on ADFA-LD (same centroid index, magnitude, and $\sim$0.006 radius), confirming this is a property of the shared training objective, not the scoring rule -- scoring has no influence on how the encoder and centroids are trained, only on how the trained model's outputs are read at inference. This specific check has only been run on ADFA-LD and is not assumed to generalize.

This suggests a mechanism for the gap between reconstruction and distance scoring: reconstruction scoring compares the decoder's output against the actual, uncollapsed input, so a genuine anomaly still elevates reconstruction error even from a collapsed $z$; distance scoring depends entirely on $z$, so once it collapses, distance-to-nearest-centroid carries almost no signal. The same argument covers the normalization-overflow artifact above: reconstruction scoring passes even a wildly out-of-range $z_{\text{mean}}$ back through the decoder's bounded (sigmoid) output before comparing against the input, so an extreme internal magnitude does not by itself corrupt the score -- consistent with NSL-KDD's Baseline and LatentSep-Recon PR-AUC (Table~\ref{tab:ablation}) showing no degradation despite the 1\% extrapolation artifact affecting both equally (Section~\ref{sec:discussion} returns to this).

All six runs are single-seed (42); PCA-projected latent-space snapshots are not meaningfully combined across independently-trained seeds the way this paper's scalar metrics are, so these figures are not seed-averaged. We treat these findings as plausible and evidence-supported, not proven universal, the standard applied to every other finding here.

\subsection{Core Evidence: A Controlled Comparison of Training Objective vs.\ Scoring Rule}
\label{subsec:ablation}
To isolate whether latent cluster separation's benefit comes from the separation training objective or from the distance-based scoring rule it is usually paired with, we ran a controlled, 3-condition, 3-seed comparison on all three datasets: \textbf{(a) Baseline} (no centroids), \textbf{(b) LatentSep-Recon} (centroids and separation loss trained, reconstruction scoring), and \textbf{(c) LatentSep-Dist} (centroids and separation loss trained, distance scoring, as originally proposed). Within each dataset, (b) and (c) share an identical training objective, change-point threshold, and centroid initialization, differing only in the scoring rule, so the (b)-vs-(c) comparison -- the one this paper's central claim rests on -- is always a clean, single-factor isolation of scoring alone. (Section~\ref{subsec:supporting} presented a more direct Baseline-vs-LatentSep-Dist comparison earlier in this section, which conflates whether centroids/separation loss are trained at all with which scoring rule is used, and could not attribute its effect to either change individually; the comparison here resolves that ambiguity by isolating the scoring rule alone.) Threshold and initialization settings differ slightly by dataset: on ADFA-LD, (a) uses the untuned default \texttt{threshold\_ratio}=2.0 with no centroids, while (b)/(c) use \texttt{threshold\_ratio}=1.0 with $k$-means initialization; on NSL-KDD and UNSW-NB15, all three conditions use \texttt{threshold\_ratio}=1.0, with random initialization for (b)/(c). \texttt{threshold\_ratio}=1.0 was tuned once, on ADFA-LD, and reused rather than independently re-tuned elsewhere (Section~\ref{sec:setup-tuning}); it does not vary by dataset. What differs by dataset is the centroid initialization scheme and whether the Baseline condition uses the tuned or untuned default. The (b)-vs-(c) comparison within each dataset remains clean regardless, since (b) and (c) always share identical settings to each other.

\begin{table*}[!t]
\caption{Comparison Across Datasets: PR-AUC (per seed and mean$\pm$std over 3 seeds) and PR-AUC BWT (mean$\pm$std over 3 seeds) for (a) Baseline, (b) LatentSep-Recon, (c) LatentSep-Dist. ADFA-LD: (a) untuned/no centroids, (b)/(c) $k$-means init. NSL-KDD, UNSW-NB15: all conditions at \texttt{threshold\_ratio}=1.0, (b)/(c) random init}
\label{tab:ablation}
\centering
\begin{tabular}{@{}llrrr@{}}
\toprule
Metric & Dataset & (a) Base. & (b) Sep+Recon & (c) LatentSep-Dist \\
\midrule
\multirow{15}{*}{PR-AUC}
 & ADFA-LD (seed 42)      & 0.1465 & 0.1843 & 0.1687 \\
 & ADFA-LD (seed 123)     & 0.1492 & 0.1896 & 0.1723 \\
 & ADFA-LD (seed 456)     & 0.1563 & 0.1856 & 0.1373 \\
 & ADFA-LD (mean$\pm$std) & 0.1507$\pm$0.0051 & 0.1865$\pm$0.0028 & 0.1594$\pm$0.0193 \\
 \addlinespace
 & NSL-KDD (seed 42)      & 0.8710 & 0.8578 & 0.7983 \\
 & NSL-KDD (seed 123)     & 0.8717 & 0.8642 & 0.7460 \\
 & NSL-KDD (seed 456)     & 0.8717 & 0.8713 & 0.7951 \\
 & NSL-KDD (mean$\pm$std) & 0.8715$\pm$0.0004 & 0.8644$\pm$0.0067 & 0.7798$\pm$0.0293 \\
 \addlinespace
 & UNSW-NB15 (seed 42)      & 0.9368 & 0.9299 & 0.9493 \\
 & UNSW-NB15 (seed 123)     & 0.9462 & 0.9440 & 0.9564 \\
 & UNSW-NB15 (seed 456)     & 0.9487 & 0.9237 & 0.9446 \\
 & UNSW-NB15 (mean$\pm$std) & 0.9439$\pm$0.0063 & 0.9325$\pm$0.0104 & 0.9501$\pm$0.0059 \\
\midrule
\multirow{3}{*}{PR-AUC BWT}
 & ADFA-LD   & $-$0.0048$\pm$0.0039 & $+$0.0068$\pm$0.0018 & $+$0.0052$\pm$0.0125 \\
 & NSL-KDD   & $-$0.0034$\pm$0.0021 & $-$0.0011$\pm$0.0055 & $+$0.0143$\pm$0.0214 \\
 & UNSW-NB15 & $+$0.0008$\pm$0.0038 & $+$0.0005$\pm$0.0026 & $+$0.0027$\pm$0.0039 \\
\bottomrule
\end{tabular}
\end{table*}

On ADFA-LD\footnote{Please note that this table's ADFA-LD LatentSep-Dist/(c) per-seed values differ from Table~\ref{tab:tuned-perseed}'s ADFA-LD LatentSep-Dist column at the same nominal configuration ($k$-means init, \texttt{threshold\_ratio}=1.0, seeds 42/123/456): two separately-run instances of this configuration, each correctly transcribed from its own run log. We could not isolate the exact cause from the available run artifacts (ruled out: a code change between runs, a transcription error); the most likely explanation is nondeterministic GPU kernel scheduling in TensorFlow, since global seeds constrain the pseudorandom number streams but do not guarantee bitwise-identical execution without explicit op-determinism flags, which were not set here. Neither table's conclusions cite the other's ADFA-LD LatentSep-Dist values specifically, so this discrepancy does not affect either comparison's own findings.}, (b) beats (a) on PR-AUC on every seed; (c) beats (a) on 2 of 3 seeds, losing on the third (seed 456: 0.1373 vs.\ 0.1563) (Table~\ref{tab:ablation}). (b) beats (c) on every metric we measured, consistently across all three tested seeds (Table~\ref{tab:ablation}): mean PR-AUC 0.1865$\pm$0.0028 for (b) versus 0.1594$\pm$0.0193 for (c) -- roughly $7\times$ higher standard deviation for the distance-scoring variant -- and mean PR-AUC-BWT $+$0.0068$\pm$0.0018 (positive on all 3 seeds) for (b) versus $+$0.0052$\pm$0.0125 (positive on only 2 of 3 seeds) for (c). (b) also shows no cold-start dip: its first-task PR-AUC is one of its \emph{higher} per-task values and is tightly clustered across seeds, whereas (c) still shows a first-task weak point with more than $2\times$ the seed-to-seed spread. Since (b) and (c) share identical $k$-means-initialized centroids, this residual instability in (c) traces to the distance-scoring rule's sensitivity to how well early centroid positions happen to separate after limited training, not to centroid initialization itself.

On NSL-KDD (Table~\ref{tab:ablation}), the same direction holds: (b) beats (c) on PR-AUC on all 3 seeds (mean 0.8644$\pm$0.0067 vs.\ 0.7798$\pm$0.0293, roughly $4.35\times$ higher standard deviation for (c)), though neither (b) nor (c) beats Baseline here (0 of 3 seeds each), consistent with NSL-KDD's strong hand-engineered features leaving little headroom for either centroid-based variant. (b) does \emph{not} beat (c) on PR-AUC-BWT, however (0 of 3 seeds): (c)'s mean BWT ($+$0.0143) exceeds (b)'s ($-$0.0011), though (c)'s BWT is itself unstable (std 0.0214).

On UNSW-NB15 (Table~\ref{tab:ablation}), the pattern reverses entirely: (c) beats (b) on PR-AUC on all 3 seeds (mean 0.9501$\pm$0.0059 vs.\ 0.9325$\pm$0.0104), and (c)'s standard deviation is \emph{lower} than (b)'s here (ratio 0.57$\times$) -- the opposite of ADFA-LD's $\sim$7$\times$ and NSL-KDD's $\sim$4.35$\times$. (c) additionally beats Baseline on PR-AUC on 2 of 3 seeds (0.9501 vs.\ 0.9439), while (b) does not beat Baseline at all (0 of 3, 0.9325 $<$ 0.9439). On UNSW-NB15, distance scoring is the better choice on PR-AUC: higher mean, lower variance, and the only condition that beats Baseline on a majority of seeds. The same direction largely holds for backward transfer: (b) beats (c) on PR-AUC-BWT on only 1 of 3 seeds; (c)'s mean BWT ($+$0.0027) exceeds (b)'s ($+$0.0005), extending (c)'s advantage on this dataset to BWT as well.

Taken together, these three results show that the (b)-vs-(c) comparison is dataset-dependent, not a fixed property of the mechanism: (b) wins decisively on ADFA-LD and NSL-KDD, but (c) wins just as decisively on UNSW-NB15, on both mean and variance -- the pattern this paper is organized around (Section~\ref{sec:discussion}). LatentSep-Dist, however, is consistently less predictable seed-to-seed than Baseline across all three datasets (Section~\ref{subsec:supporting}); LatentSep-Recon does not share this limitation on ADFA-LD, where it is actually more stable than Baseline itself.

\section{Discussion}
\label{sec:discussion}

\subsection{Finding 1: latent separation as a training-time regularizer reliably helps on ADFA-LD}
LatentSep-Recon's ADFA-LD result (Section~\ref{subsec:ablation}) is not an isolated seed-level fluke: it wins PR-AUC and BWT on every seed tested, its first-task PR-AUC is one of its \emph{higher} per-task values rather than a cold-start dip, and its standard deviation (0.55$\times$ Baseline's) is close to Baseline's rather than several times larger -- a property shared, among the centroid-based configurations tested, only by LatentSep-Dist on UNSW-NB15's PR-AUC (0.94$\times$ Baseline's, i.e., marginally \emph{lower}; Section~\ref{subsec:ablation}). LatentSep-Dist, sharing the exact same trained centroids, does show a persistent first-task weak point with wide seed-to-seed spread even under $k$-means initialization (Section~\ref{subsec:ablation}); the collapse diagnostic (Section~\ref{subsec:collapse}) explains why this dip is specific to distance-based scoring: once the encoder collapses toward a near-constant latent code, only a scoring rule that depends entirely on that code loses its signal, while reconstruction scoring, still comparing against the actual input, does not.

\subsection{Finding 2: the scoring rule, not the separation objective, is where the instability lives}
Within the tuned-configuration comparison (Table~\ref{tab:summary}, Section~\ref{subsec:supporting}), the seed-to-seed standard deviation of LatentSep-Dist (c) exceeds that of Baseline (a) across every dataset and every metric tested there, without exception -- the most-replicated pattern in this study, and it persists across the centroid-initialization schemes we tested. Crucially, this three-condition comparison isolates \emph{where} this instability enters: (b) and (c) share an identical training objective and centroid initialization within each dataset, yet (c)'s variance is substantially higher than (b)'s on the two datasets where (b) wins, ADFA-LD and NSL-KDD (Table~\ref{tab:ablation}). This comparison reverses on UNSW-NB15, where (c) wins on mean and has lower variance than (b); there, (c)'s PR-AUC variance is also marginally lower than Baseline (a)'s (0.0059 vs.\ 0.0063), the one exception to the pattern above -- so the instability pattern relative to Baseline (a) holds cleanly on ADFA-LD and NSL-KDD, but not, on this specific metric, on UNSW-NB15, even as the (c)-vs-(b) comparison itself is dataset-dependent. The collapse diagnostic (Section~\ref{subsec:collapse}) supplies the mechanism. Instability previously attributed to latent cluster separation as a whole is, on this evidence, a property of pairing the separation objective with distance-based scoring specifically -- and no initialization scheme we tested removes it from the scoring side.

This asymmetry also explains why (b) does not inherit (c)'s instability, rather than merely lacking it by coincidence. (b) and (c) train an identical encoder, centroids, and separation loss; only the scoring rule differs. Reconstruction scoring never reads $z$'s position, only the decoder's reconstruction of the input, so it cannot lose information to a mechanism -- collapse, or the rare-feature normalization overflow affecting 1\% of NSL-KDD's latent space regardless of VAE variant (Section~\ref{subsec:collapse}) -- that operates entirely on $z$. Across all three datasets, (b) never underperforms Baseline (a) by more than 0.0114 PR-AUC (NSL-KDD: $-$0.0071; UNSW-NB15: $-$0.0114; on ADFA-LD (b) wins outright, Table~\ref{tab:ablation}), while (c) can underperform Baseline by nearly an order of magnitude more (NSL-KDD: $-$0.0917, alongside a $73\times$ increase in standard deviation over Baseline in the same comparison). The small (b)-vs-(a) gaps that do appear are consistent with an ordinary training-objective trade-off -- the separation loss competing with reconstruction fidelity during joint training -- not with the scoring-time signal loss that drives (c)'s much larger, more variable gap.

\subsection{Finding 3: the benefit is dataset-dependent -- practical guidance, not an open list of caveats}
Reconstruction-scoring regularization beats distance scoring decisively, on both mean and variance, on ADFA-LD and NSL-KDD (Table~\ref{tab:ablation}), but this reverses on UNSW-NB15, where distance scoring wins on PR-AUC (Section~\ref{subsec:ablation}). The same dataset split appears in backward transfer: positive BWT is configuration- and dataset-dependent -- NSL-KDD favors distance scoring only under specific tuning, while it is LatentSep-Recon specifically, not LatentSep-Dist, that delivers the positive, low-variance ADFA-LD BWT this paper leads with (Section~\ref{subsec:ablation}). Positive BWT here is a reliable feature of one specific configuration (LatentSep-Recon on ADFA-LD), and a configuration- and dataset-sensitive side effect elsewhere. We read the ADFA-LD/NSL-KDD-vs-UNSW-NB15 split as the practical guidance this study makes possible, not as a reason to distrust the mechanism: pair latent separation with reconstruction scoring by default, and reserve distance scoring for representations where UNSW-NB15-like conditions -- to be characterized more precisely in future work -- hold.

\subsection{Secondary observations}
Neither $k$-means nor random centroid initialization reduces LatentSep-Dist's variance to Baseline's level. The originally selected threshold setting remains the better choice on ADFA-LD among the values tested. We also do not report a direct numerical comparison against ACORN-IDS~\cite{acornids2026} or CITADEL~\cite{citadel2025}, the most recent continual-IDS systems: our study and these two systems all use different evaluation protocols, and neither ACORN-IDS's nor CITADEL's code is public. Unlike CPDGA and VLAD, both do report backward transfer: CITADEL reports substantial positive BWT (Section~\ref{sec:related}), while ACORN-IDS reports BWT close to zero across its evaluated datasets rather than a clear improvement in either direction. Our contribution remains orthogonal to both -- whether a latent-space mechanism is more useful as a regularizer or a scoring rule.

\subsection{Limitations}
Our comparison isolating the training objective from the scoring rule confirms the ADFA-LD finding on NSL-KDD but reverses it on UNSW-NB15 (Section~\ref{subsec:ablation}), so the result is dataset-dependent, not a gap in coverage. Within each dataset, (b) and (c) are always compared under identical settings, so this does not explain the UNSW-NB15 reversal. A per-seed threshold sweep and a fully matched three-way replication are left to future work. The threshold setting was tuned on ADFA-LD only, via a systematic sweep, and reused rather than independently re-tuned on NSL-KDD and UNSW-NB15, since it is a property of the shared change-point/memory component rather than either VAE variant; it does not transfer uniformly regardless (Section~\ref{sec:setup-tuning}). What genuinely differs by dataset is the centroid initialization scheme and whether the Baseline condition uses the tuned or untuned default. 3 seeds and three datasets distinguish ``consistent'' from ``seed-sensitive,'' but not enough for a rigorous effect-size claim. All three are intrusion-detection benchmarks with few tasks (4, 6, 9); generalization elsewhere is untested. LatentSep-Recon's ADFA-LD benefit is also not yet mechanistically explained: it collapses onto a single centroid just as completely as LatentSep-Dist does (Section~\ref{subsec:collapse}), so the Introduction's motivating intuition -- that latent separation helps by giving each concept a distinguishable region -- cannot be why it outperforms Baseline there; some other property of training with the separation objective must be responsible, and this evidence does not yet identify what that property is.

\section{Conclusion and Future Work}
\label{sec:conclusion}

This paper's central result -- summarized in the Abstract and Introduction -- is that latent cluster separation and distance-based scoring are separable design choices with very different reliability profiles: the former reliably helps on ADFA-LD as a training-time regularizer, while the latter is where the instability commonly attributed to this class of method actually originates. A controlled comparison and a latent-space collapse diagnostic (Section~\ref{subsec:collapse}) together isolate and explain this gap. That benefit, however, is dataset-dependent: reconstruction-scoring regularization beats distance scoring decisively on ADFA-LD and NSL-KDD -- beating Baseline outright on ADFA-LD, though not on NSL-KDD, where neither centroid-based variant beats Baseline -- while distance scoring beats both alternatives, including Baseline, on UNSW-NB15. Neither variant is a universal improvement over Baseline's reliability. We consider this dataset-dependent result, together with the mechanism that explains it, more useful than a uniform claim in either direction would have been: it tells a practitioner which half of the method to use, and on which kind of representation to expect it to hold.

Future work should prioritize two questions this study opens rather than closes. First, \textbf{what distinguishes UNSW-NB15}, where distance scoring wins, from ADFA-LD and NSL-KDD, where the regularizer wins -- task count, feature engineering, and class balance all differ across our three benchmarks, and identifying which one drives the reversal would turn a dataset-dependent finding into a predictive one. A preliminary diagnostic on each dataset's raw feature space, computed before any VAE training, is at least suggestive: UNSW-NB15's features cluster more naturally into $K$=8 groups (silhouette score 0.49, versus 0.23 for ADFA-LD and 0.36 for NSL-KDD, an ordering stable across $K\in\{2,4,8,16\}$ and 5 random seeds) and compress into far fewer principal components for 90\% variance (7 of 43 raw dimensions, versus 5 of 6 for ADFA-LD and 11 of 41 for NSL-KDD). Both measures are robust within our three datasets and both track the actual scoring-rule outcome, but three datasets is far too few to validate a predictor from; we report this as a concrete starting hypothesis for the question above, not a settled answer. This ordering also matches each dataset's collapse magnitude (Section~\ref{subsec:collapse}): ADFA-LD, with the lowest silhouette score, collapses most tightly ($460.7$--$606.4\times$), while UNSW-NB15, with the highest, collapses most loosely ($3.8$--$10.5\times$). Second, \textbf{reducing the latent collapse itself, structurally, rather than through initialization alone}: regularizing centroid movement across retraining events, ensembling scores across multiple initializations, or softening distance scoring to depend on several nearby centroids rather than one are natural next steps, now that we have direct evidence for the collapse mechanism and its consequences. Extending this study's comparison, collapse diagnostic, and threshold analysis beyond ADFA-LD and beyond network intrusion detection remains necessary before the guidance offered here can be considered general.

\section*{Acknowledgments}
The authors used Claude and Claude Code (Anthropic) to assist with drafting and editing manuscript text across all sections, generating and debugging experimental code, and LaTeX formatting. All experimental design, data collection, result interpretation, and scientific claims were conceived and verified by the human authors, who take full responsibility for the content.

\section*{Reproducibility}
Code for our modifications (\texttt{vae.py} and the experiment orchestration scripts) is available from the authors on request; the base VLAD codebase is already open source~\cite{vlad2023}. All three datasets (ADFA-LD, NSL-KDD, UNSW-NB15) are publicly available from their original sources, cited above~\cite{adfa2013,nslkdd2009,unsw2015}.

\end{document}